\documentclass{article}
\usepackage{spconf,amsmath,graphicx,hyperref}

\usepackage[dvipsnames]{xcolor}
\usepackage{enumitem}

\title{AbracADDbra: Touch-Guided Object Addition\\ by Decoupling Placement and Editing Subtasks}
\name{Kunal Swami, Raghu Chittersu, Yuvraj Rathore, Rajeev Irny, Shashavali Doodekula, Alok Shukla}
\address{Samsung Research India Bangalore\\
\{kunal.swami, raghu.c, y.rathore, rajeev.i, shasha.d, alok.shukla\}@samsung.com}
\begin{document}
%
\maketitle
\begin{abstract}
Instruction-based object addition is often hindered by the ambiguity of text-only prompts or the tedious nature of mask-based inputs. To address this usability gap, we introduce \emph{AbracADDbra}, a user-friendly framework that leverages intuitive touch priors to spatially ground succinct instructions for precise placement. Our efficient, decoupled architecture uses a vision-language transformer for touch-guided placement, followed by a diffusion model that jointly generates the object and an instance mask for high-fidelity blending. To facilitate standardized evaluation, we contribute the Touch2Add benchmark for this interactive task. Our extensive evaluations, where our placement model significantly outperforms both random placement and general-purpose VLM baselines, confirm the framework's ability to produce high-fidelity edits. Furthermore, our analysis reveals a strong correlation between initial placement accuracy and final edit quality, validating our decoupled approach. This work thus paves the way for more accessible and efficient creative tools.
\end{abstract}
\begin{keywords}
Instruction-based Object Addition, Touch-guided Object Addition, Generative Models
\end{keywords}
\section{Introduction}
\label{sec:intro}

\begin{figure}[t!]
	\centering
	\includegraphics[scale=0.44]{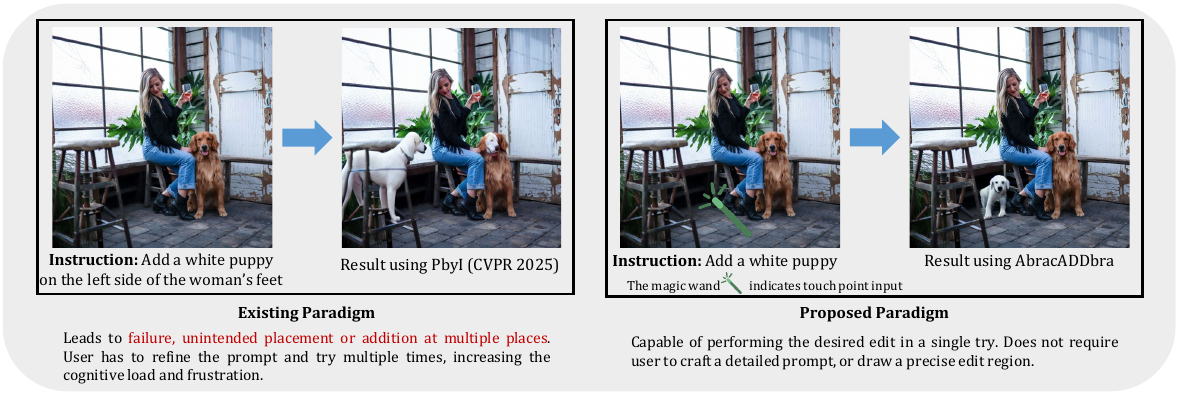}
	\vspace{-0.5cm}
	\caption{\textbf{Main idea of \emph{AbracADDbra}.} We propose a new framework for adding objects to images that lets users provide simple touch input along with instructions, making editing more accurate and user-friendly.}
	\label{fig:ournewparadigm} 
	\vspace{-0.5cm}
\end{figure}

\begin{figure}
	\includegraphics[scale=0.435]{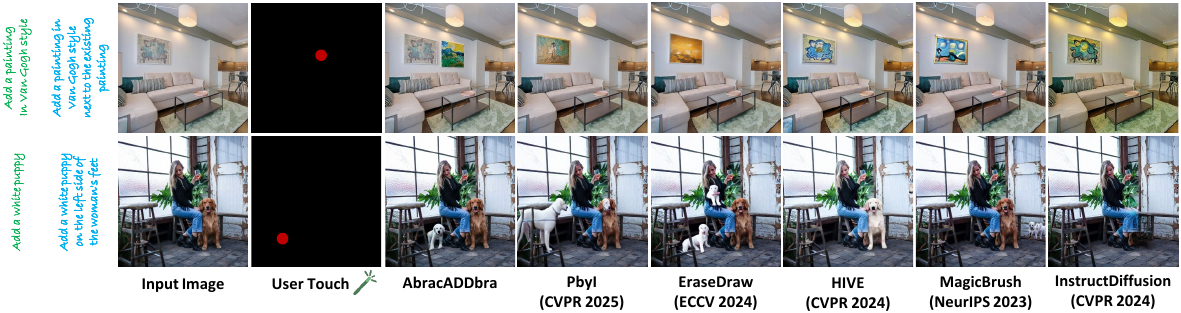}
    \vspace{-0.8cm}
	\caption{\emph{AbracADDbra} performs high-fidelity object addition via touch and succinct instructions. Our method combines an intuitive touch prior with a simple, succinct prompt (in \textcolor{ForestGreen}{green}) to achieve precise object addition. For a fair comparison against strong baselines, we provided them with detailed prompts (in \textcolor{Cyan}{blue}).}
	\label{fig:teaser}
	\vspace{-0.4cm}
\end{figure}

Our work challenges the prevailing paradigm in instruction-based object addition, where methods either rely on cumbersome masks or struggle to interpret complex textual instructions accurately. We argue that existing approaches overlook the potential of intuitive visual prompts for guiding object placement, thereby defining a new problem in this domain. Modern user interfaces, with their emphasis on touch and click-based interactions, offer a natural and intuitive way to provide such visual guidance. To address this, we introduce a novel framework that seamlessly integrates textual instructions with user touch priors, pioneering a new paradigm for flexible and interactive object addition (see Fig.~\ref{fig:ournewparadigm}). This approach empowers users to provide precise guidance with minimal effort, streamlining the editing process and significantly reducing the cognitive burden. For instance, instead of crafting detailed instructions like ``add a bird on the right side of the bench opposite to the crow," a user can simply say ``add a bird" and provide a touch point for the desired location. This deceptively simple yet powerful approach yields a remarkable outcome (see Fig.~\ref{fig:teaser} ). Our key contributions include the efficient, decoupled framework itself, comprehensive evaluations validating its superiority in placement accuracy over strong baselines, and a demonstration of state-of-the-art editing fidelity. This work establishes touch-guidance as a highly effective paradigm for user-centric generative editing.

To support research and standardized evaluation of this new paradigm, we introduce our benchmark, Touch2Add. This dataset comprises $544$ carefully curated image, add instruction, and user touch triplets, encompassing diverse scenarios and enabling a thorough assessment of model capabilities for this task.

Following are the \textbf{major contributions} of this work:
\vspace{-0.24cm}
\begin{itemize}[leftmargin=*]
	\setlength{\itemsep}{0cm}
	\item A novel, touch-guided paradigm for instruction-based object addition, enabled by a framework that decouples placement and editing. This provides fine-grained control and a more intuitive user experience compared to traditional mask-based or text-only approaches.
	\item The introduction of Touch2Add, a new benchmark dataset with $544$ diverse triplets (image, instruction, touch point) to facilitate standardized evaluation of this interactive editing task.
	\item Comprehensive validation on multiple benchmarks that highlights our method's advantages: achieving superior edit quality over text-only methods while enabling a more intuitive and efficient user experience through touch interaction.
\end{itemize}

\section{Related Work}
\label{sec:relatedwork}
\vspace{-0.2cm}

\subsection{Instruction-based Image Editing.} 
\vspace{-0.1cm}
Instruction-based image editing has garnered significant attention, with approaches broadly categorized into mask-free \cite{ip2p_cvpr2023,magicbrush_neurips2023,instructdiffusion_cvpr2024,hive_cvpr2024} and mask-based methods \cite{glide_icml2022,blended_latent_diffusion_siggraph2023,smartbrush_cvpr2023}. InstructPix2Pix (IP2P) \cite{ip2p_cvpr2023} pioneered this domain by training a diffusion model on a large-scale dataset of paired images and textual instructions. Subsequent works like MagicBrush \cite{magicbrush_neurips2023} and HIVE \cite{hive_cvpr2024} built upon IP2P by extending training datasets and incorporating human feedback respectively. Alternatively, some methods \cite{plugnplay_diffusion_cvpr2023} explore training-free editing using techniques like null-text inversion \cite{nulltextinv_editing_cvpr2023} and attention map manipulation. SmartEdit \cite{smartedit_cvpr2024} and MGIE \cite{mgie_iclr2024} integrate Multimodal Large Language Models (MLLMs) to perform image editing that requires understanding and reasoning complex instructions. However, these methods often necessitate elaborate textual prompts to achieve the desired edits, increasing the cognitive burden on users and hindering their practical usability.

\vspace{-0.25cm}
\subsection{Instruction-based Object Addition.} 
\vspace{-0.1cm}
The challenge of instruction-based object addition has recently garnered increased attention \cite{pbyi_cvpr2025,erasedraw_eccv2024}. These works propose methods for creating large-scale datasets by artificially removing objects from existing image segmentation datasets \cite{cocodataset_cvpr2018} using inpainting techniques \cite{lama_inpainting_wacv2022,stablediffusion_cvpr2022}. Subsequently, a text-to-image diffusion UNet \cite{stablediffusion_cvpr2022} is fine-tuned on this data to perform object addition. In contrast to these approaches, which rely solely on textual instructions, we introduce a novel framework that incorporates intuitive user interaction to minimize the cognitive burden and enhance precision. Furthermore, we explicitly decouple the task into distinct placement and editing subtasks, enabling specialized learning and optimization for improved performance.

\section{Proposed Method}
\label{sec:proposedmethod}
\vspace{-0.1cm}

\subsection{Problem Formulation}
\label{subsec:problemformulation}
\vspace{-0.15cm}
Given an input image $\mathcal{I}$, an object addition instruction $\mathcal{T}$, and a user touch-point $(\tau_x, \tau_y)$ in the image coordinate space, our objective is to add an object at the specified location while adhering to the instruction. We decompose this task into two distinct subtasks: placement and editing. The placement subtask involves predicting the location and scale of the object to be added, utilizing $\mathcal{I}$, $\mathcal{T}$, and $(\tau_x, \tau_y)$ as input to a placement prediction model. The editing subtask employs a diffusion-based editing model to perform instruction-based object addition, leveraging $\mathcal{I}$, $\mathcal{T}$, and the output of the placement prediction model.

\begin{figure}[t!]
	\centering
	\includegraphics[scale=0.21]{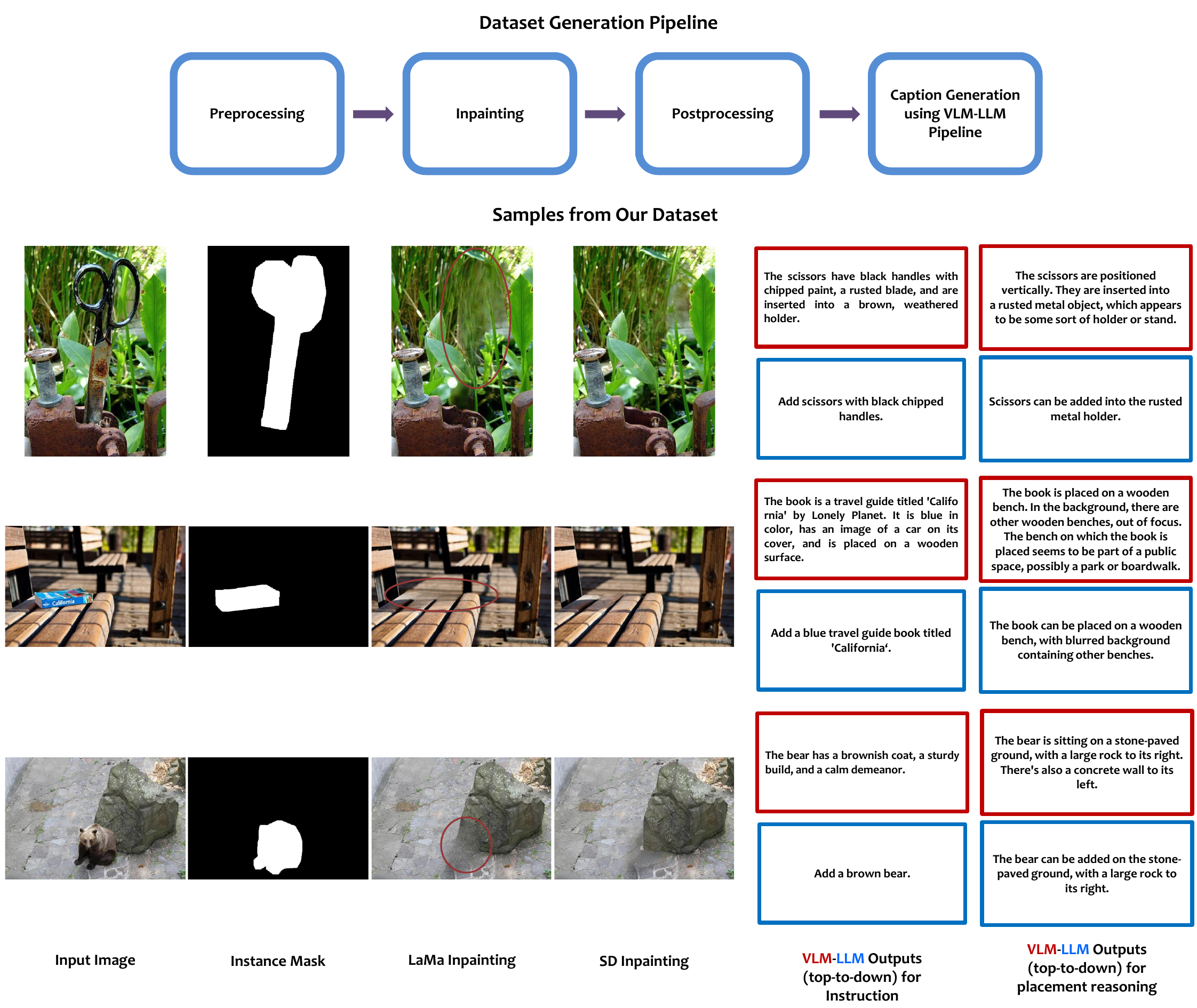}
	\vspace{-0.45cm}
	\caption{Our automated data generation pipeline. The process includes four stages: (1) Preprocessing: Filtering COCO objects based on size, boundary proximity, and CLIP score. (2) Inpainting: A two-stage process using LaMa \cite{lama_inpainting_wacv2022} and Stable Diffusion \cite{stablediffusion_cvpr2022}. (3) Postprocessing: Applying filtering steps from \cite{pbyi_cvpr2025}. (4) Caption Generation: Using GLaMM \cite{glamm_cvpr2024} and an LLM to create the instruction and placement reasoning.}
	\label{fig:datasetgenerationpipeline} 
	\vspace{-0.6cm}
\end{figure}

\vspace{-0.3cm}
\subsection{Training Dataset Generation} 
\label{subsec:datasetconstruction}
\vspace{-0.15cm}
To train our models for this novel task, we generate a large-scale dataset of image-instruction quadruplets, which includes target images and textual \emph{placement reasoning} captions. Our automated pipeline, illustrated in Fig.~\ref{fig:datasetgenerationpipeline}, is designed to produce high-quality and diverse training data. Starting with images from COCO \cite{cocodataset_cvpr2018}, we generate target images using a hybrid inpainting approach that leverages the complementary strengths of LaMa \cite{lama_inpainting_wacv2022} for robust context preservation and Stable Diffusion \cite{stablediffusion_cvpr2022} for high-fidelity detail synthesis. Rich textual instructions and reasoning captions are then derived using GLaMM \cite{glamm_cvpr2024} and an LLM. This process yields a dataset of $100,000$ samples, which we split into $90,000$ for training and $10,000$ for validation.

\subsection{Placement Prediction Model}
\label{subsec:placementpredictionmodel}

\subsubsection{Architecture}
\begin{figure}[t!]
	\centering
	\includegraphics[scale=0.33]{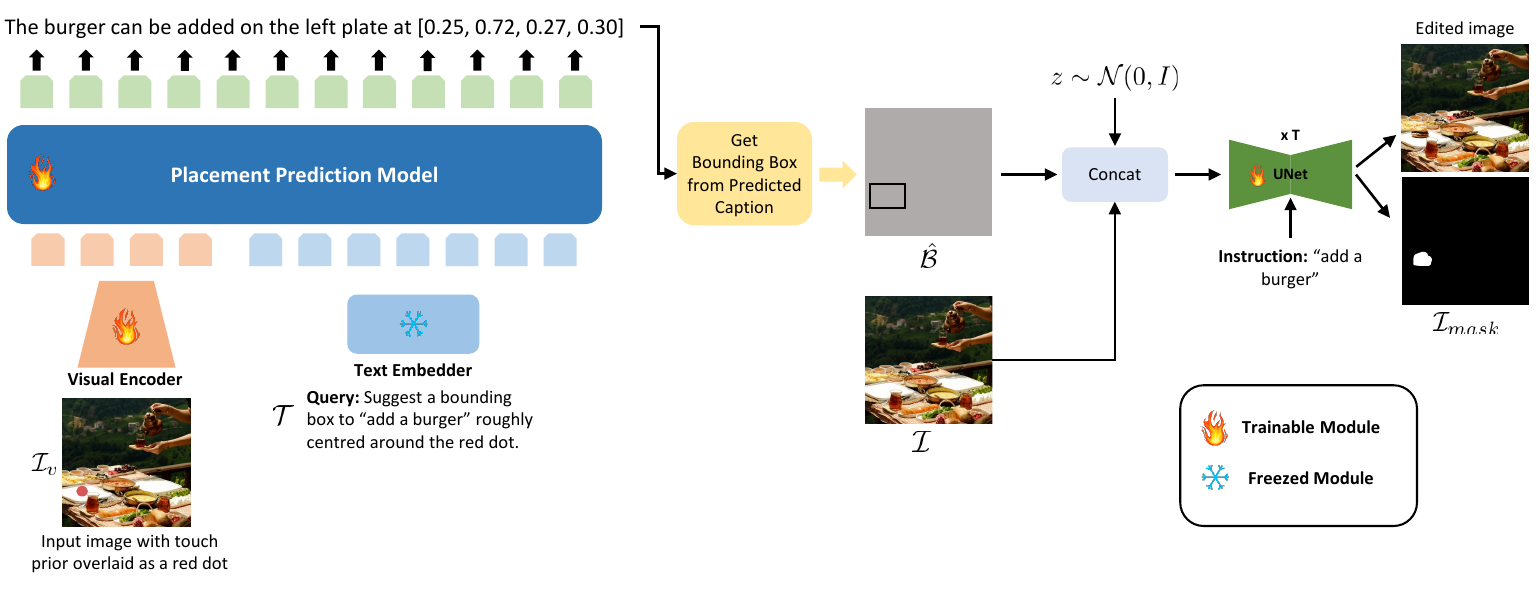}
	\vspace{-0.9cm}
	\caption{The detailed architecture of our method. Inference scenario is shown and VAE encoder and decoder are omitted.}
	\label{fig:modelarchitecture} 
	\vspace{-0.3cm}
\end{figure}

As illustrated in Fig. ~\ref{fig:modelarchitecture}, our placement model is a lightweight Vision-Language Model (VLM) that integrates a pre-trained vision encoder ($V_{enc}$) with a GPT-$2$ Large \cite{gpt2_arxiv2019} decoder ($LLM_{dec}$). To enable the decoder to jointly reason over visual and textual information, we pre-train this VLM using the LLaVA \cite{llava_neurips2023} methodology. We selected GPT-$2$ Large to effectively balance the need for robust language comprehension with the computational efficiency required for an interactive system.

\vspace{-0.2cm}
\subsubsection{Input Formulation}
To condition the model on both user instruction and spatial intent, we formulate the inputs as follows. First, we directly embed the user's touch prior into the visual stream. The touch point $\mathcal{P} = (\tau_x, \tau_y)$ is rendered as a visual marker (a $10 \times 10$ red dot) onto the input image $\mathcal{I}$ using alpha blending ($\alpha=0.4$) to create a composite image $\mathcal{I}_{v}$. This approach is both effective and computationally efficient \cite{viplava_cvpr2024}. The vision encoder then processes this composite image, $F_v = V_{enc}(\mathcal{I}_{v})$. Second, we prompt the model with a textual query $\mathcal{T}$ that leverages the visual marker: \texttt{Suggest a bounding box to <add object instruction> roughly centered at the red dot}.

\vspace{-0.2cm}
\subsubsection{Output Formulation and Training Objective}
We formulate placement prediction as a dual-objective task, training the model to autoregressively generate both a textual \emph{placement reasoning} caption and the corresponding normalized bounding box coordinates $B = [x_c, y_c, w, h]$. We hypothesize that generating explicit reasoning enhances model understanding, a claim supported by our ablations. Following recent work \cite{pink_mllm_cvpr2024}, the bounding box coordinates are quantized and treated as a sequence of text tokens, allowing the entire output to be generated by the $LLM_{dec}$.

To enhance robustness to inexact user touches, we augment the ground-truth centroid $(x_c, y_c)$ with random perturbations: $x'_c = x_c + \delta(w)$ and $y'_c = y_c + \delta(h)$, where $\delta(p) \sim \text{Clamp}_{[-1, 1]}(\mathcal{N}(0, 1/3)) \cdot p/2$. The ground-truth response $R_{gt}$ is a concatenation of the reasoning caption and these augmented, tokenized coordinates. The model is then trained end-to-end by minimizing a standard autoregressive language modeling loss ($\mathcal{L}_{LM}$):

\vspace{-0.1cm}
\begin{equation}
	\mathcal{L}_{LM} = -\sum_{t=1}^{T} \log p(R_t | R_{<t}, F_v, \mathcal{T})
	\label{eq:llmloss}
\end{equation}
\vspace{-0.1cm}

where $p(R_t | \dots)$ is the probability of predicting the token $R_t$ conditioned on previous tokens and the multimodal inputs.

\vspace{-0.2cm}
\subsection{Instruction-based Object Addition Model}
\label{subsec:objectadditionmodel}

Our editing model is built upon a Stable Diffusion v$1.5$ \cite{stablediffusion_cvpr2022} backbone. We adapt its UNet to jointly predict the edited image and a corresponding instance mask \cite{smartbrush_cvpr2023}, a process crucial for achieving high-fidelity blending and preserving unedited background regions. The model is conditioned on the input image, the textual instruction, and the predicted placement bounding box. The training objective is a weighted sum of a standard denoising loss for the image and a Dice loss \cite{dice_loss_miccai2017} for the instance mask.

\vspace{-0.2cm}
\section{Results and Discussion}
\label{sec:experiments}

\vspace{-0.1cm}
\subsection{Evaluation Protocol}
\begin{figure}[t!]
	\centering
	\includegraphics[scale=0.213]{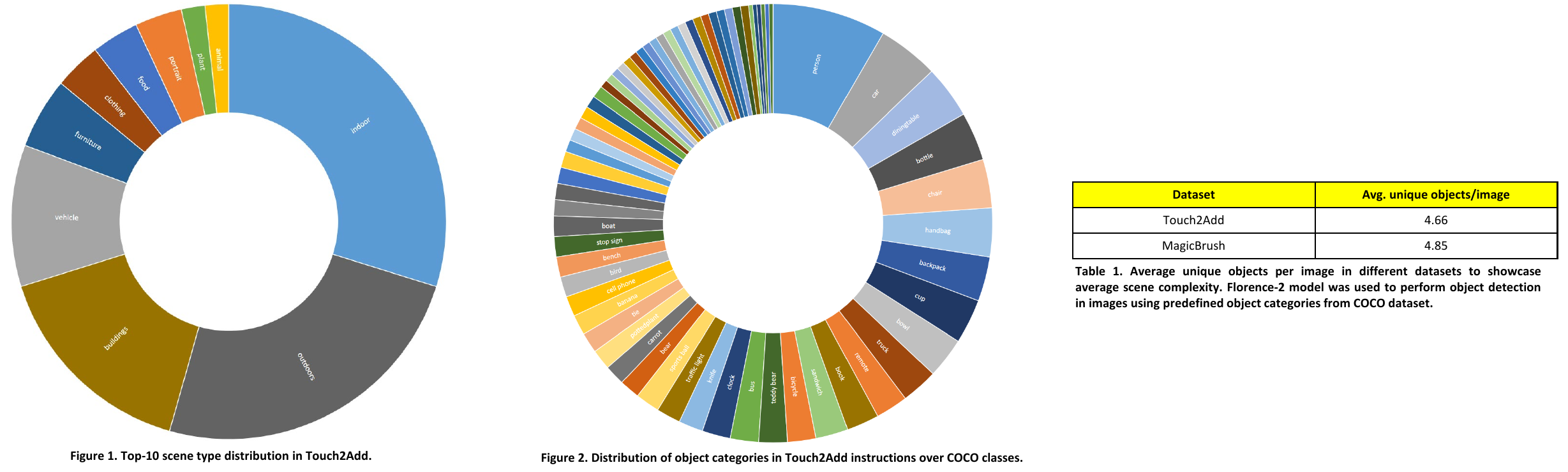}
	\vspace{-0.8cm}
	\caption{Diversity statistics of our Touch2Add dataset. We also compare the average scene complexity with the MagicBrush object addition subset.}
	\label{fig:touch2addstats} 
	\vspace{-0.3cm}
\end{figure}

We conduct a comprehensive evaluation on our Touch2Add benchmark (see dataset statistics in Fig.~\ref{fig:touch2addstats}) and the $150$ object addition subset of the MagicBrush \cite{magicbrush_neurips2023} dataset. Our protocol combines quantitative fidelity metrics ($L_1$, $L_2$, CLIP, DINO) with two distinct user studies for a holistic assessment. The first study required participants to rank the top three outputs based on overall edit quality. The second study focused specifically on the placement subtask, where participants ranked outputs based on placement plausibility and scale appropriateness, \emph{with the original touch point hidden to prevent bias}.

We compare against several state-of-the-art object addition \cite{pbyi_cvpr2025,erasedraw_eccv2024} and image editing \cite{ip2p_cvpr2023,magicbrush_neurips2023,instructdiffusion_cvpr2024,hive_cvpr2024} methods. To specifically isolate the contribution of our learned placement model, we also introduce a strong \emph{Random Placement} baseline. In this baseline, bounding boxes are randomly generated around the user's touch point, with width and height sampled from a Gaussian distribution derived from training data statistics. It provides a robust lower bound for placement accuracy.

\vspace{-0.1cm}
\subsection{Main Results}
\label{subsec:mainresults}

\begin{figure}[t!]
	\centering
	\includegraphics[scale=0.42]{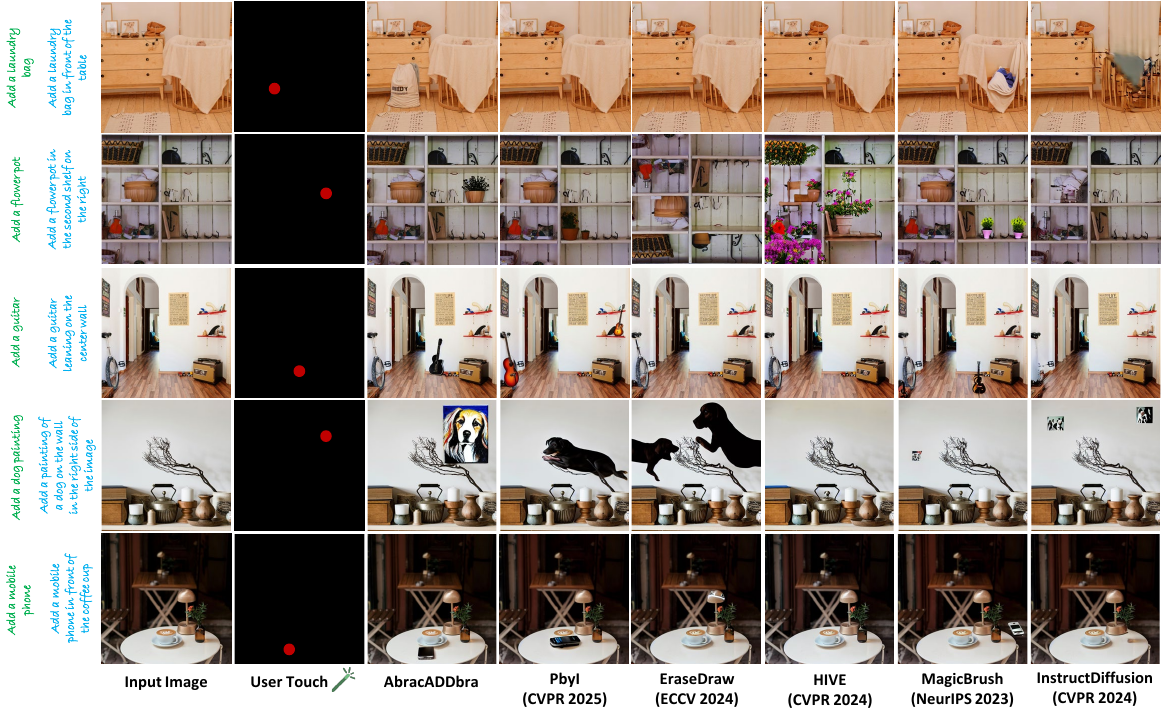}
	\vspace{-0.45cm}
	\caption{Qualitative comparison on the Touch2Add benchmark, where \emph{AbracADDbra} uses succinct prompts (in \textcolor{ForestGreen}{green}) while baselines receive more detailed prompts (in \textcolor{Cyan}{blue}) for a fair evaluation.}
	\label{fig:qualitativecomparison} 
	\vspace{-0.5cm}
\end{figure}

Our method's effectiveness is rooted in its highly accurate placement model. On our Touch2Add benchmark, it achieves a mean IoU approximately $52$\% higher than the \emph{Random Placement} baseline (see Tab.~\ref{tab:quantitativecomparison_placement_touch2add}). This quantitative superiority is also confirmed by our user study results in the same table, where \emph{AbracADDbra} was ranked first or second for placement quality in the majority of cases. This precise placement directly translates to superior editing fidelity, as shown qualitatively (Fig.~\ref{fig:qualitativecomparison}) and quantitatively (Tab.~\ref{tab:touch2add_quantitativecomparison_quality}, \ref{tab:magicbrush_quantitativecomparison_quality}). Our full framework excels on both Touch2Add and MagicBrush benchmarks.

Our experiments provide compelling evidence for our core architectural hypotheses. \textbf{First}, the surprisingly competitive performance of the \emph{Random Placement} baseline on final editing metrics reveals a critical weakness in existing methods: they struggle with accurate localization, and providing \emph{any} spatial prior is highly beneficial. This observation strongly supports our hypothesis that decoupling placement and editing is crucial for achieving high-fidelity object addition. \textbf{Second}, we demonstrate the necessity of specialized training; our dedicated placement model significantly outperforms powerful, general-purpose VLMs like LLaVA \cite{llava_neurips2023} and ViP-LLaVA \cite{viplava_cvpr2024} (Tab.~\ref{tab:quantitativecomparison_placementbaselines}), proving that tailored training is essential for this specialized task.

\begin{table}[t!]
	\centering
	\caption{Placement assessment in instruction-based object addition task performed on \textbf{Touch2Add benchmark} dataset. `-' means not applicable.}
	\resizebox{0.48\textwidth}{!}{
		\begin{tabular}{|c|ccccc|}
			\hline
			\textbf{Methods} & \textbf{mean IoU$\uparrow$} & \textbf{IoU $> 0.5$ (\%) $\uparrow$} & \textbf{U.S. Rank-1 (\%)$\uparrow$} & \textbf{U.S. Rank-2 (\%)$\uparrow$} & \textbf{U.S. Rank-3 (\%)$\uparrow$} 	\\ \hline
			PbyI \cite{pbyi_cvpr2025}                				& -				& -				& 10.61	       			& 10.88  				& 12.20 									\\ 
			EraseDraw \cite{erasedraw_eccv2024}       				& -				& -				& 13.00	       			& 10.61 				& \underline{15.65}							\\ \hline
			IP2P \cite{ip2p_cvpr2023}		          				& -				& -				& 5.57         			& 8.75 					& 13.00 									\\
			MagicBrush \cite{magicbrush_neurips2023}  				& -				& -				& 6.10         			& 9.28 					& \underline{15.65} 						\\
			InstructDiffusion \cite{instructdiffusion_cvpr2024}		& -     		& -				& 3.18         			& 1.33 					& 4.51 										\\
			HIVE \cite{hive_cvpr2024}		          				& -				& -				& 3.45         			& 6.37 					& 8.49 										\\ \hline
			Random Placement  					      				& 0.31			& 0.56			& \underline{20.95}     & \underline{17.24} 	& \textbf{17.77} 							\\ \hline
			\emph{AbracADDbra}                              	    & \textbf{0.47}	& \textbf{0.84}	& \textbf{37.14}        & \textbf{35.54} 		& 12.73 									\\ \hline
		\end{tabular}
	}
	\label{tab:quantitativecomparison_placement_touch2add}
	\vspace{-0.6cm}
\end{table}

\begin{table}[t!]
	\centering
	\caption{Quantitative comparison for instruction-based object addition on the \textbf{Touch2Add benchmark} dataset. U. S. stands for user study.}
	\resizebox{0.48\textwidth}{!}{
		\begin{tabular}{|c|ccccccc|}
			\hline
			\textbf{Methods} & \textbf{CLIP $\uparrow$} & \textbf{DINO $\uparrow$}  & \textbf{L1$\downarrow$} & \textbf{L2$\downarrow$} & \textbf{U.S. Rank-1 (\%)$\uparrow$} & \textbf{U.S. Rank-2 (\%)$\uparrow$} & \textbf{U.S. Rank-3 (\%)$\uparrow$}\\ \hline
			PbyI \cite{pbyi_cvpr2025}              				& 0.9143	   			& 0.9122  				& \underline{0.0319}  & 0.0061				& 9.28					& 10.61				& 11.94					\\ 
			EraseDraw \cite{erasedraw_eccv2024}         		& 0.9135	   			& 0.9279 			    & 0.0350      		  & 0.0079				& 13.00					& 9.81				& \textbf{18.30}		\\ \hline
			IP2P \cite{ip2p_cvpr2023}		       				& 0.8157       			& 0.7830 			    & 0.0710      		  & 0.0160     			& 4.24				    & 9.28				& 10.88					\\
			MagicBrush \cite{magicbrush_neurips2023}			& 0.8898       			& 0.9099 				& 0.0376      		  & 0.0079     	    	& 5.84				    & 8.49				& \underline{16.45}		\\
			InstructDiffusion \cite{instructdiffusion_cvpr2024} & \underline{0.9259} 	& \underline{0.9378}  	& 0.0339      		  & \underline{0.0055}  & 2.65					& 2.12				& 5.31					\\
			HIVE \cite{hive_cvpr2024}		       	     	    & 0.8971       			& 0.8991 				& 0.0440      		  & 0.0067     			& 4.77					& 6.63				& 7.43					\\ \hline
			Random placement							 		& 0.9146       			& 0.8851 				& 0.0355      		  & 0.0077     			& \underline{17.51}		& \underline{23.34}	& \underline{16.45}		\\ \hline
			\emph{AbracADDbra}  								& \textbf{0.9304}	    & \textbf{0.9821} 		& \textbf{0.0264}     & \textbf{0.0020}     & \textbf{42.71}    	& \textbf{29.71}	& 13.26					\\ \hline
		\end{tabular}
	}
	\label{tab:touch2add_quantitativecomparison_quality}
	\vspace{-0.55cm}
\end{table}

\begin{table}[t!]
	\centering
	\caption{Quantitative comparison for instruction-based object addition on the \textbf{MagicBrush benchmark} dataset. U. S. stands for user study.}
	\resizebox{0.48\textwidth}{!}{
		\begin{tabular}{|c|ccccccc|}
			\hline
			\textbf{Methods} & \textbf{CLIP $\uparrow$} & \textbf{DINO $\uparrow$}  & \textbf{L1$\downarrow$} & \textbf{L2$\downarrow$} & \textbf{U.S. Rank-1 (\%)$\uparrow$} & \textbf{U.S. Rank-2 (\%)$\uparrow$} & \textbf{U.S. Rank-3 (\%)$\uparrow$}\\ \hline
			PbyI \cite{pbyi_cvpr2025}            					& 0.8715	  			& 0.8922  			& 0.0411  				& 0.0103				& 10.95					& 18.10					& 15.71				\\ 
			EraseDraw \cite{erasedraw_eccv2024}        				& 0.8882	   			& 0.9191 			& 0.0406      			& 0.0096				& 12.38					& 13.81					& 10.00				\\ \hline
			IP2P \cite{ip2p_cvpr2023}		    					& 0.7854       			& 0.7762 			& 0.0815      			& 0.0192     			& 3.81					& 3.33					& 9.05				\\
			MagicBrush \cite{magicbrush_neurips2023}				& 0.8533       			& 0.9060			& 0.0458      			& 0.0121     			& \underline{13.33}		& \underline{19.05}		& \textbf{25.24}	\\ 
			InstructDiffusion \cite{instructdiffusion_cvpr2024} 	& \underline{0.8956} 	& \textbf{0.9366}  	& \underline{0.0373}    & \underline{0.0070}  	& 5.71					& 18.57					& \underline{20.00}	\\
			HIVE \cite{hive_cvpr2024}		       	     	    	& 0.8593       			& 0.8686 			& 0.0535      		  	& 0.0092     			& 0.48					& 3.33					& 10.00				\\ \hline
			\emph{AbracADDbra}  									& \textbf{0.9312}	    & \underline{0.9330}& \textbf{0.0119}     	& \textbf{0.0052}     	& \textbf{53.33}    	& \textbf{23.81}	    & 10.00				\\ \hline
		\end{tabular}
	}
	\label{tab:magicbrush_quantitativecomparison_quality}
	\vspace{-0.6cm}
\end{table}

\begin{table}[t!]
	\centering
	\caption{Placement model comparison against baselines.}
	\resizebox{0.35\textwidth}{!}{
		\begin{tabular}{|c|cc|}
			\hline
			\textbf{Methods} & \textbf{mean IoU$\uparrow$} & \textbf{IoU $> 0.5$ (\%) $\uparrow$}	        \\ \hline
			LLaVA \cite{llava_neurips2023}		      & 0.09			& 0.02	 			    			\\
			ViP-LLaVA \cite{viplava_cvpr2024}		  & 0.12			& 0.14								\\ \hline
			\emph{AbracADDbra} Placement Model        & \textbf{0.47}	& \textbf{0.84} 					\\ \hline
		\end{tabular}
	}
	\label{tab:quantitativecomparison_placementbaselines}
	\vspace{-0.21cm}
\end{table}

\begin{figure}[t!]
	\centering
	\includegraphics[scale=0.35]{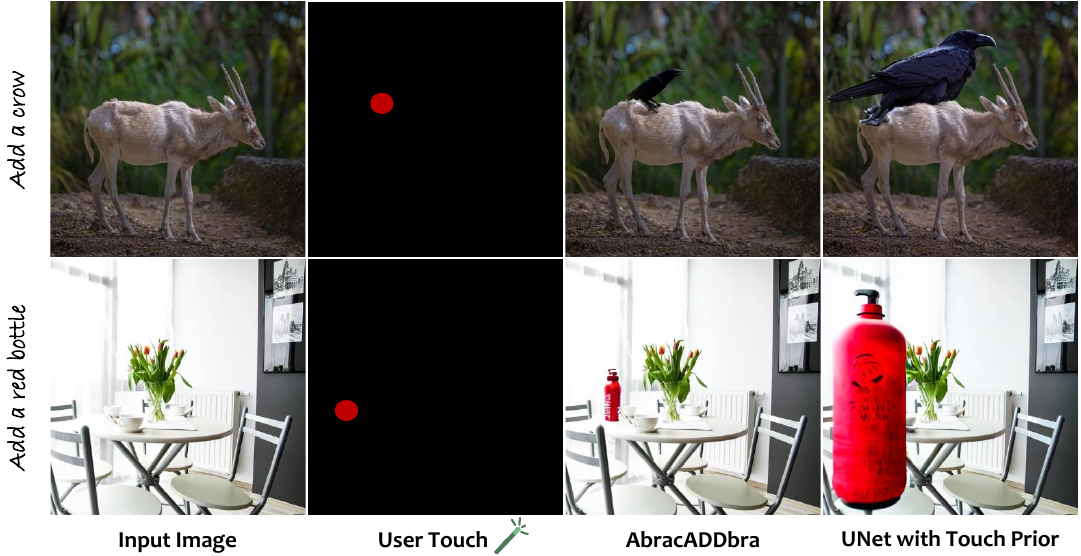}
	\vspace{-0.45cm}
	\caption{Ablation study. Impact of dedicated placement model.}
	\label{fig:ablationstudy_impactofdedicatedplacementmodel} 
	\vspace{-0.57cm}
\end{figure}

\begin{table}[t!]
	\centering
	\caption{Ablation study. Need of placement model.}
	\resizebox{0.45\textwidth}{!}{
		\begin{tabular}{|c|ccccc|}
			\hline
			\textbf{Methods} 		            & \textbf{CLIP$\uparrow$} 		& \textbf{DINO$\uparrow$} 	& \textbf{L1$\downarrow$}  		& \textbf{L2$\downarrow$} 			& \textbf{Win Rate (\%)$\uparrow$}		\\ \hline
			UNet with Touch Prior			    & 0.9110		              	& 0.9589 	 			    & 0.0370      		  			& 0.0075                            & 39.67                                 \\
			\emph{AbracADDbra}                  & \textbf{0.9304}		        & \textbf{0.9821} 		    & \textbf{0.0264}     			& \textbf{0.0020}                   & \textbf{60.33}                        \\ \hline
		\end{tabular}
	}
	\label{tab:ablationstudy_unetusertouchprior}
	\vspace{-0.6cm}
\end{table}

\begin{table}[t!]
	\centering
	\caption{Ablation study. Impact of placement reasoning.}
	\resizebox{0.35\textwidth}{!}{
		\begin{tabular}{|c|cc|}
			\hline
			\textbf{Methods} 		            & \textbf{mean IoU$\uparrow$} & \textbf{IoU $> 0.5$ (\%) $\uparrow$}			\\ \hline
			W/o Reasoning Loss			        & 0.44		                  & 0.81 	 			    	    				\\
			W/ Reasoning Loss                   & \textbf{0.47}		          & \textbf{0.84} 						    		\\ \hline
		\end{tabular}
	}
	\label{tab:ablationstudy_placementmodelwithoutreasoning}
	\vspace{-0.6cm}
\end{table}

\vspace{-0.2cm}
\subsection{Ablation Studies}
\label{subsec:ablationstudies}
\vspace{-0.2cm}

To validate our decoupled design, we ablate the dedicated placement model and instead condition the diffusion UNet directly on a visual touch prior (a $12\times12$ mask). As shown in Tab.~\ref{tab:ablationstudy_unetusertouchprior}, this simplified approach is significantly outperformed by our full \emph{AbracADDbra} model, which was also preferred by users in a head-to-head study ($60$\% preference). Qualitatively (Fig.~\ref{fig:ablationstudy_impactofdedicatedplacementmodel}), while this baseline can localize the edit, it consistently fails to estimate the appropriate object scale, resulting in contextually inconsistent outputs. This confirms that a dedicated placement model is crucial for predicting both location and scale.

We also ablate the placement reasoning objective from our placement model's training (see Sec.~\ref{subsec:placementpredictionmodel}). As shown in Tab.~\ref{tab:ablationstudy_placementmodelwithoutreasoning}, removing this task results in a notable drop in IoU. This result supports our hypothesis that training the model to generate explicit reasoning improves its final placement accuracy.

\vspace{-0.35cm}
\section{Conclusion and Future Work}
\label{sec:conclusion}
\vspace{-0.2cm}
We introduced \emph{AbracADDbra}, a novel framework that enhances instruction-based object addition by incorporating intuitive user touch priors within an efficient, decoupled architecture. Our comprehensive evaluations on our new Touch2Add benchmark and the MagicBrush dataset demonstrate that our approach achieves superior quality and instruction alignment over text-only methods by effectively resolving spatial ambiguity. A key insight from our analysis is the strong correlation between initial placement accuracy and final edit quality, which validates our decoupled design. We contribute the Touch2Add benchmark to foster standardized research in this interactive domain. This work validates touch interaction as a highly effective paradigm, our future work will generalize this approach to a broader range of edits.


\bibliographystyle{IEEEbib}
\bibliography{references}

@String(CVPR= {IEEE Conf. Comput. Vis. Pattern Recog.})

@String(ECCV= {Eur. Conf. Comput. Vis.})

@String(ICLR = {Int. Conf. Learn. Represent.})

@String(CVPR  = {CVPR})

@String(ECCV  = {ECCV})

@String(ICLR  = {ICLR})

@inproceedings{pbyi_cvpr2025,
  author       = {Navve Wasserman and
                  Noam Rotstein and
                  Roy Ganz and
                  Ron Kimmel},
  title        = {Paint by Inpaint: Learning to Add Image Objects by Removing Them First},
  booktitle	   = {{CVPR}},
  year         = {2025},
}

@inproceedings{erasedraw_eccv2024,
  author={Canberk, Alper and Bondarenko, Maksym and Ozguroglu, Ege and Liu, Ruoshi and Vondrick, Carl},
  title={EraseDraw: Learning to Draw Step-by-Step via Erasing Objects from Images},
  booktitle={{ECCV}},
  year={2024}
}

@inproceedings{hive_cvpr2024,
  author       = {Shu Zhang and
                  Xinyi Yang and
                  Yihao Feng and
                  Can Qin and
                  Chia{-}Chih Chen and
                  Ning Yu and
                  Zeyuan Chen and
                  Huan Wang and
                  Silvio Savarese and
                  Stefano Ermon and
                  Caiming Xiong and
                  Ran Xu},
  title        = {{HIVE:} Harnessing Human Feedback for Instructional Visual Editing},
  booktitle    = {{CVPR}},
  pages        = {9026--9036},
  year         = {2024},
}

@inproceedings{instructdiffusion_cvpr2024,
  author       = {Zigang Geng and
                  Binxin Yang and
                  Tiankai Hang and
                  Chen Li and
                  Shuyang Gu and
                  Ting Zhang and
                  Jianmin Bao and
                  Zheng Zhang and
                  Houqiang Li and
                  Han Hu and
                  Dong Chen and
                  Baining Guo},
  title        = {InstructDiffusion: {A} Generalist Modeling Interface for Vision Tasks},
  booktitle    = {{CVPR}},
  pages        = {12709--12720},
  year         = {2024},
}

@inproceedings{magicbrush_neurips2023,
	author       = {Kai Zhang and
	Lingbo Mo and
	Wenhu Chen and
	Huan Sun and
	Yu Su},
	title        = {MagicBrush: {A} Manually Annotated Dataset for Instruction-Guided Image Editing},
	booktitle    = {{NeurIPS}},
	year         = {2023},
}

@inproceedings{ip2p_cvpr2023,
	author       = {Tim Brooks and
	Aleksander Holynski and
	Alexei A. Efros},
	title        = {InstructPix2Pix: Learning to Follow Image Editing Instructions},
	booktitle    = {{CVPR}},
	pages        = {18392--18402},
	year         = {2023},
}

@article{blended_latent_diffusion_siggraph2023,
	author       = {Omri Avrahami and
	Ohad Fried and
	Dani Lischinski},
	title        = {Blended Latent Diffusion},
	journal      = {{ACM} Trans. Graph.},
	volume       = {42},
	number       = {4},
	pages        = {149:1--149:11},
	year         = {2023},
}

@inproceedings{smartbrush_cvpr2023,
	author       = {Shaoan Xie and
	Zhifei Zhang and
	Zhe Lin and
	Tobias Hinz and
	Kun Zhang},
	title        = {SmartBrush: Text and Shape Guided Object Inpainting with Diffusion Model},
	booktitle    = {{CVPR}},
	pages        = {22428--22437},
	year         = {2023},
}

@inproceedings{glide_icml2022,
	author       = {Alexander Quinn Nichol and
	Prafulla Dhariwal and
	Aditya Ramesh and
	Pranav Shyam and
	Pamela Mishkin and
	Bob McGrew and
	Ilya Sutskever and
	Mark Chen},
	title        = {{GLIDE:} Towards Photorealistic Image Generation and Editing with Text-Guided Diffusion Models},
	booktitle    = {{ICML}},
	pages        = {16784--16804},
	year         = {2022},
}

@inproceedings{smartedit_cvpr2024,
  author       = {Yuzhou Huang and
                  Liangbin Xie and
                  Xintao Wang and
                  Ziyang Yuan and
                  Xiaodong Cun and
                  Yixiao Ge and
                  Jiantao Zhou and
                  Chao Dong and
                  Rui Huang and
                  Ruimao Zhang and
                  Ying Shan},
  title        = {SmartEdit: Exploring Complex Instruction-Based Image Editing with
                  Multimodal Large Language Models},
  booktitle    = {{CVPR}},
  pages        = {8362--8371},
  year         = {2024},
}

@inproceedings{mgie_iclr2024,
  author       = {Tsu{-}Jui Fu and
                  Wenze Hu and
                  Xianzhi Du and
                  William Yang Wang and
                  Yinfei Yang and
                  Zhe Gan},
  title        = {Guiding Instruction-based Image Editing via Multimodal Large Language Models},
  booktitle    = {{ICLR}},
  year         = {2024},
}

@inproceedings{plugnplay_diffusion_cvpr2023,
	author       = {Narek Tumanyan and
	Michal Geyer and
	Shai Bagon and
	Tali Dekel},
	title        = {Plug-and-Play Diffusion Features for Text-Driven Image-to-Image Translation},
	booktitle    = {{CVPR}},
	pages        = {1921--1930},
	year         = {2023},
}

@inproceedings{stablediffusion_cvpr2022,
	author       = {Robin Rombach and
	Andreas Blattmann and
	Dominik Lorenz and
	Patrick Esser and
	Bj{\"{o}}rn Ommer},
	title        = {High-Resolution Image Synthesis with Latent Diffusion Models},
	booktitle    = {{CVPR}},
	pages        = {10674--10685},
	year         = {2022},
}

@inproceedings{nulltextinv_editing_cvpr2023,
	author       = {Ron Mokady and
	Amir Hertz and
	Kfir Aberman and
	Yael Pritch and
	Daniel Cohen{-}Or},
	title        = {Null-text Inversion for Editing Real Images using Guided Diffusion Models},
	booktitle    = {{CVPR}},
	pages        = {6038--6047},
	year         = {2023},
}

@inproceedings{glamm_cvpr2024,
  author       = {Hanoona Abdul Rasheed and
                  Muhammad Maaz and
                  Sahal Shaji Mullappilly and
                  Abdelrahman M. Shaker and
                  Salman H. Khan and
                  Hisham Cholakkal and
                  Rao Muhammad Anwer and
                  Eric P. Xing and
                  Ming{-}Hsuan Yang and
                  Fahad Shahbaz Khan},
  title        = {GLaMM: Pixel Grounding Large Multimodal Model},
  booktitle    = {{CVPR}},
  pages        = {13009--13018},
  year         = {2024},
}

@inproceedings{pink_mllm_cvpr2024,
  author       = {Shiyu Xuan and
                  Qingpei Guo and
                  Ming Yang and
                  Shiliang Zhang},
  title        = {Pink: Unveiling the Power of Referential Comprehension for Multi-modal
                  LLMs},
  booktitle    = {{CVPR}},
  pages        = {13838--13848},
  year         = {2024},
}

@inproceedings{viplava_cvpr2024,
  author       = {Mu Cai and
                  Haotian Liu and
                  Siva Karthik Mustikovela and
                  Gregory P. Meyer and
                  Yuning Chai and
                  Dennis Park and
                  Yong Jae Lee},
  title        = {ViP-LLaVA: Making Large Multimodal Models Understand Arbitrary Visual Prompts},
  booktitle    = {{CVPR}},
  pages        = {12914--12923},
  year         = {2024},
}

@inproceedings{llava_neurips2023,
    author      = {Liu, Haotian and Li, Chunyuan and Wu, Qingyang and Lee, Yong Jae},
    title       = {Visual Instruction Tuning},
    booktitle   = {{NeurIPS}},
    year        = {2023}
}

@article{gpt2_arxiv2019,
  title={Language models are unsupervised multitask learners},
  author={Radford, Alec and Wu, Jeffrey and Child, Rewon and Luan, David and Amodei, Dario and Sutskever, Ilya and others},
  journal={OpenAI blog},
  volume={1},
  number={8},
  pages={9},
  year={2019}
}

@inproceedings{cocodataset_cvpr2018,
	author       = {Holger Caesar and
	Jasper R. R. Uijlings and
	Vittorio Ferrari},
	title        = {COCO-Stuff: Thing and Stuff Classes in Context},
	booktitle    = {{CVPR}},
	pages        = {1209--1218},
	year         = {2018},
}

@inproceedings{lama_inpainting_wacv2022,
  author       = {Roman Suvorov and
                  Elizaveta Logacheva and
                  Anton Mashikhin and
                  Anastasia Remizova and
                  Arsenii Ashukha and
                  Aleksei Silvestrov and
                  Naejin Kong and
                  Harshith Goka and
                  Kiwoong Park and
                  Victor Lempitsky},
  title        = {Resolution-robust Large Mask Inpainting with Fourier Convolutions},
  booktitle    = {{WACV}},
  pages        = {3172--3182},
  year         = {2022},
}

@inproceedings{dice_loss_miccai2017,
  author       = {Carole H. Sudre and
                  Wenqi Li and
                  Tom Vercauteren and
                  S{\'{e}}bastien Ourselin and
                  M. Jorge Cardoso},
  title        = {Generalised Dice Overlap as a Deep Learning Loss Function for Highly
                  Unbalanced Segmentations},
  booktitle    = {{MICCAI}},
  pages        = {240--248},
  year         = {2017},
}

\end{document}